\documentclass[runningheads]{llncs}
\usepackage{graphicx}
\usepackage{tikz}
\usepackage{comment}
\usepackage{hhline}
\usepackage{hyperref}
\usepackage{amsmath,amsfonts,amssymb}% define this before the line numbering.
\usepackage{color}
\usepackage{arydshln}
\usepackage[accsupp]{axessibility}  % Improves PDF readability for those with disabilities.

\begin{document}
	% \renewcommand\makeLineNumber {\hss\thelinenumber\ {6mm} \rlap{\hskip\textwidth\ \hspace{6.5mm}\thelinenumber}}
	% \linenumbers
	\pagestyle{headings}
	\mainmatter
	\def\ECCVSubNumber{74}  % Insert your submission number here	
	\title{Multi-Scale Attention-based Multiple Instance Learning for Classification of Multi-Gigapixel Histology Images} % Replace with your title	
	\begin{comment}
		\titlerunning{ECCV-22 submission ID \ECCVSubNumber} 
		\authorrunning{ECCV-22 submission ID \ECCVSubNumber} 
		\author{Anonymous ECCV submission}
		\institute{Paper ID \ECCVSubNumber}
	\end{comment}
	%******************
	
	%\begin{comment}
	\titlerunning{Multi-Scale Attention-based Multiple Instance Learning}
	\author{
		Made Satria Wibawa\inst{1} \and
		Kwok-Wai Lo\inst{2}\and
		Lawrence S. Young\inst{3}\and
		Nasir Rajpoot\inst{1,4}
	}
	\authorrunning{M.S. Wibawa et al.}
	\institute{
		Tissue Image Analytics Centre, Department of Computer Science, University of Warwick \\
		\email{\{made-satria.wibawa,n.m.rajpoot\}@warwick.ac.uk}\\ \and 
		Department of Anatomical and Cellular Pathology, The Chinese University of Hong Kong \\
		\email{kwlo@cuhk.edu.hk} \\ \and
		Warwick Medical School, University of Warwick \\
		\email{l.s.young@warwick.ac.uk} \\ \and
		The Alan Turing Institute, London}
	%\end{comment}
	%******************
	\maketitle
	\setcounter{footnote}{0}	
	
	\begin{abstract}
	Histology images with multi-gigapixel of resolution yield rich information for cancer diagnosis and prognosis. Most of the time, only slide-level label is available because pixel-wise annotation is labour intensive task. In this paper, we propose a deep learning pipeline for classification in histology images. Using multiple instance learning, we attempt to predict the latent membrane protein 1 (LMP1) status of nasopharyngeal carcinoma (NPC) based on haematoxylin and eosin-stain (H\&E) histology images. We utilised attention mechanism with residual connection for our aggregation layers. In our 3-fold cross-validation experiment, we achieved average accuracy, AUC and F1-score 0.936, 0.995 and 0.862, respectively. This method also allows us to examine the model interpretability by visualising attention scores. To the best of our knowledge, this is the first attempt to predict LMP1 status on NPC using deep learning.
		
	\end{abstract}

	\section{Introduction}
	Analysing histology images for cancer diagnosis and the prognosis is not a task without challenges. A Whole Slide Image (WSI) in \(40\times\) magnification with an average resolution of \(200,000\times150,000\) contains 30 billion pixels and a size of \(\sim\)90GB in the uncompressed state. Moreover, histology image mainly has a noisy/ambiguous label. Most of the time label of the WSI is assigned on the slide level, for example, the label for cancer stage. This may create ambiguous interpretations by the machine learning/deep learning model, because in such a large region, not all regions correspond to the slide label. Some regions may contain tumours, and the rest is non-tumour regions e.g. background, stroma, and lymphocyte.

	The standard approach to handling multi-gigapixel WSIs is by extracting its region into smaller patches that the machine can process. The noisy label problem also occurs in this approach since we do not have a label for each patch. The multiple instance learning (MIL) paradigm is generally used to overcome this problem \cite{CARBONNEAU2018329}. In MIL, each patch is represented as an instance in a bag. Since WSIs have more than one patch, the bag contains multiple instances, hence the name 'multiple' instances learning. During training, only global (slide-level) image labels are required for supervision. Aggregation mechanism is then utilised to summarise all information in instances to make a final prediction.
	
	The selection of aggregation mechanism is one of the vital parameter in MIL. Several aggregation mechanisms have been introduced in MIL for computational histopathology, such as the max and the median score \cite{klein2021deep} or the average of all patches scores \cite{coudray2018classification,schaumberg2018h}. The limitation of such mechanisms is they are not trainable. Therefore, attention mechanism \cite{bahdanau2014neural} was utilised in the recent MIL model on computational pathology \cite{qiu2021attention,ilse2018attention}. Attention mechanism is trainable, it computes the weights of the instances representation. Strong weights/scores indicate instances are more important to final prediction than instances with low scores.
	
	Several studies have already utilised MIL for cancer diagnosis and prognosis. For example, Lu and colleagues \cite{lu2021data} proposed attention-based multiple instance learning for subtyping renal cell carcinoma and non-small-cell lung carcinoma. Campanella and colleagues \cite{campanella2019clinical} used multiple instance learning with RNN-based aggregation to discriminate tumour regions in breast, skin, and prostate cancer. No current study examines the use of multiple instance learning in nasopharyngeal carcinoma.
	
	Nasopharyngeal carcinoma (NPC) is a malignancy that develops from epithelial cells within lymphocyte-rich nasopharyngeal mucosa. The high incidence rate of NPC mainly occurs in southern China, southeast Asia and north Africa, which are 50-100 times greater than the rates in other regions of the world. NPC has unique pathogenesis involving genetic, lifestyle and viral (Epstein-Barr virus or EBV) cofactors \cite{wong2021nasopharyngeal}. EBV infection on NPC encodes several viral oncoprotein, one of the most important ones is latent membrane protein (LMP1). LMP1 is the predominant oncogenic driver of NPC \cite{kieser2015latent}. Overexpression of LMP1 promotes NPC progression by inducing invasive growth of human epithelial and nasopharyngeal cells. Due its nature, LMP1 is an excellent therapeutic target for EBV-associated NPC \cite{lee2019nasopharyngeal}. However, detection of LMP1 on NPC with standard testing such as immunohistochemistry may cause additional costs and delays the diagnosis. On the other hand, histopathology images are widely available and provide rich information of the cancer on the tissue level. Deep learning-based algorithms have been applied for several tasks and deliver promising results in EBV-related cancer in histopathology. Deep learning has been used to predict EBV status in gastric cancer \cite{zheng2022deep}, predict EBV integration sites \cite{Liang2021} and predict microsatellite instability status in gastric cancer \cite{muti2021development}.	However, despite the advance in computational histopathology and the significant importance of LMP1 in NPC progression, this topic is still an understudy area.
	
	In this paper, we proposed a deep learning pipeline based on the MIL paradigm to predict LMP1 status in NPC. Our main contributions in this paper are: (1)To the best of our knowledge, this is the first study that attempts to utilise deep learning on LMP1 status prediction based on histology images of NPC. (2)We propose multi-scale attention-based with residual connection\footnote{Model code: \href{https://github.com/mdsatria/MultiAttentionMIL}{https://github.com/mdsatria/MultiAttentionMIL}} for aggregation layer in MIL. (3)Our proposed MIL pipeline outperformed other known MIL methods for predicting LMP1 status in NPC.

	% \section{Related Works}
	% In last couple

	\section{Materials and Methods}
	\subsection{Dataset}
	This study used two kinds of datasets: the first for the tissue classification task and the second is the LMP1 dataset. 
	\begin{itemize}
		\item \textbf{Tissue classification dataset}. For this task, we combined two datasets from Kather100K \cite{kather2019predicting} dataset and an internal dataset. Kather100K contains 100,000 non-overlapping image patches from H\&E stained histological images of colorectal cancer. The number of tissue classes in this dataset is nine, we exclude the normal colon mucosa class (NORM) from the dataset. We argued that the NORM class is irrelevant in head and neck tissue. Thus, only eight classes from the Kather100K dataset were utilised in this study. The internal dataset consists of three cohorts. The number of images patches in this dataset is 180,344 with seven classes which taken at 20x magnification. We combined images with the same classes from Kather100K and internal datasets in the final dataset. 
		
		\item \textbf{LMP1 dataset}. LMP1 data was collected from an University Hospital. The dataset comprises 101 NPC cases with a H\&E stained whole slide images (WSIs) for each case. All WSIs were taken by an Aperio scanner at 40x magnification and 0.250 microns per pixel resolution. Expression of LMP1 was determined by immunohistochemical staining. The proportion score was according to the percentage of tumour cells with positive membrane and cytoplasmic staining (0–100). The intensity score was assigned for the average intensity of positive tumour cells (0,none; 1,weak; 2,intermediate; 3,strong). The LMP1 staining score was the product of proportion and intensity scores, ranging from 0 to 300. The LMP1 expression was categorized into absence/low/negative (score 0–100) and high/positive (score 101–300). Details of the class distribution of the dataset in this study can be seen in Table \ref{tab:table_dataset}. %The detailed protocol regarding the determination of LMP1 expression can be found in here \cite{li2017exome}.% 
	
		\setlength{\tabcolsep}{4pt}
		\begin{table}
			\begin{center}
				\caption{Details of the dataset}
				\label{tab:table_dataset}
				\begin{tabular}{l c}
					\hline\noalign{\smallskip}
					Label & Number of Cases	\\  
					\noalign{\smallskip}
					\hline
					\noalign{\smallskip}
					LMP1 positive & 25 \\
					LMP1 negative & 76 \\
				
					\hline
					\noalign{\smallskip}
					Total cases & 101 \\ 
					\hline
				\end{tabular}
			\end{center}
		\end{table}
	\setlength{\tabcolsep}{1.4pt}
		
	\end{itemize}

	\subsection{Tissue Classification}
	We utilised a pretrained ResNet-50 \cite{he2016deep} with ImageNet \cite{deng2009imagenet} to classify images into tissue and non-tissue classes. Dataset for tissue classification was split into three parts, i.e. training, validation and testing with the data distribution of 80\%, 10\% and 10\%, respectively. Training images were augmented in several methods, including colour augmentation, kernel filter (sharpening and blurring) and geometric transformation. Before images used in training, we normalised the value of the pixel within the range of -1 and 1.
	
	We trained all layers in the ResNet50 model, using an Adam optimizer with starting learning rate of \(1\times{10}^{-3}\) and weight decay of \(5\times{10}^{-4}\). Training was conducted on 20 epochs and every 10 epochs, we decreased learning rate by ten times. To regularize the model, we apply a dropout layer with the probability of 0.5 before the fully connected layer. We use cross entropy to calculate the loss of the network which is defined as:
	
	\begin{equation}
		CE=\sum_{i=1}^C t_{i}\log(f(s)_i)
	\end{equation}
	where
	\begin{equation}
		f(s)_i = \frac{e^s_i}{\sum_{j=1}^{C}e^s_j} 
	\end{equation}
	\(t_i\) is the ground truth label and \(s_i\) is the predicted label and \(f(s)_i\) is a softmax activation function. In our case, there were eight classes, thus \(C\in [0..7] \). 
		
	\subsection{LMP1 Prediction}
\begin{figure}%[ht]
		\centering
		\includegraphics[width=\textwidth]{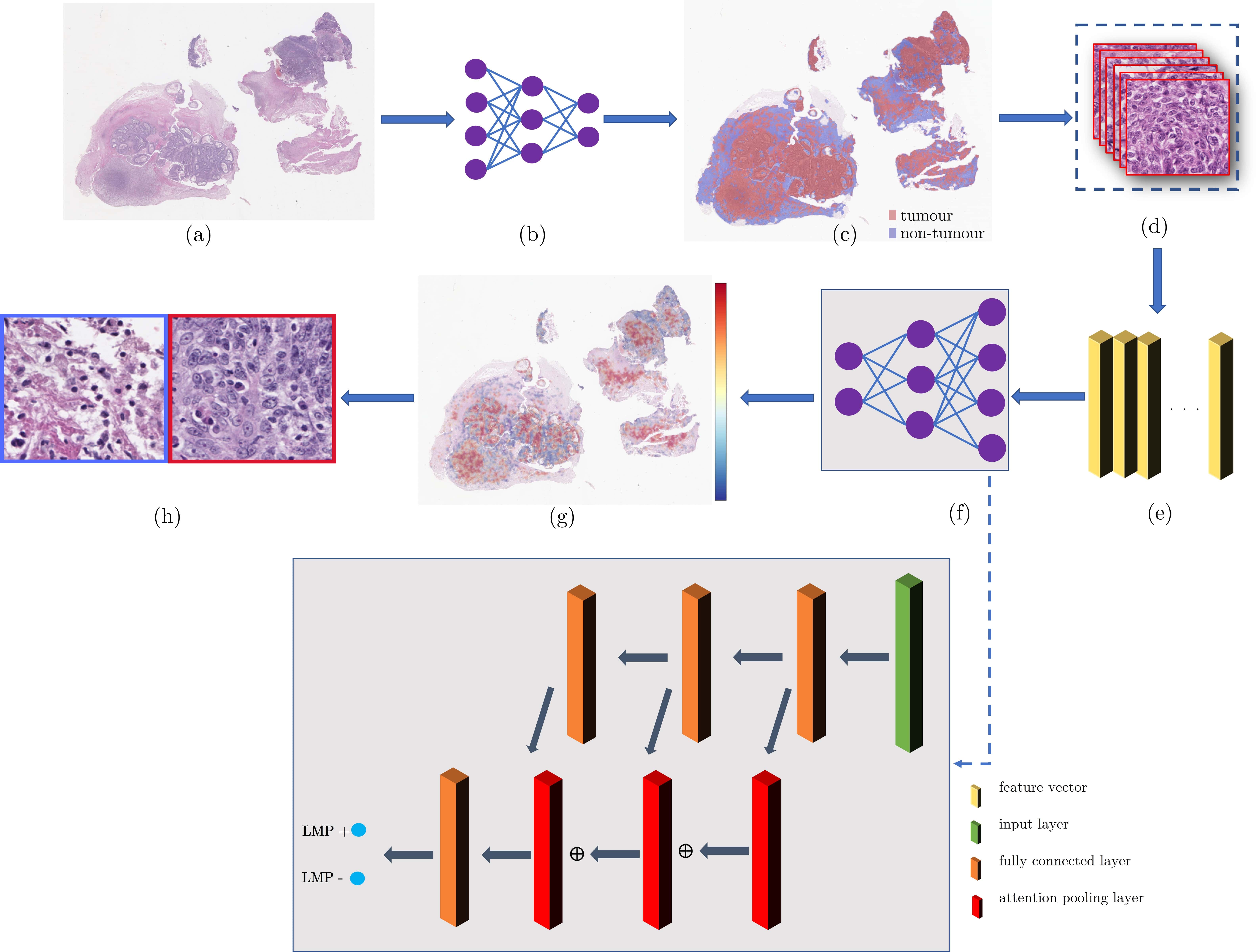}
		\caption{\textbf{The workflow of LMP1 Prediction.} Patches from WSIs (a) were classified into the tumour and non-tumour regions by the tumour classification model based on ResNet50 (b). The tumour regions of WSIs can be seen as the red area of the heatmap (c). All tumour patches were then used as instances of the bag (d) and encoded into feature vectors (e) with backbone networks. Our MIL model which consists of four fully connected layers, and three attention-based aggregation layers with the residual connection (f) then trained with these feature vectors. The model will generate attention scores for each patch. The red area in the heatmap shows a high attention score (g\&h).} \label{fig_methods}
	\end{figure}

	The main pipeline of LMP1 prediction is illustrated in Figure \ref{fig_methods}. After tumour patches were identified in the tissue classification stage, we encoded all the tumour patches into feature vectors. We experimented with two backbone networks for encoding patches into feature vectors, namely ResNet-18 and EfficientNet-B0 \cite{tan2019efficientnet}.
	
	Let \(B\) denote the bag, \(x\) is the feature vector extracted from backbone network and \(K\) is the number of patches in the slide, bag of instances will be \(B = \{x_1, ..., x_K\}\). Within the ResNet-18 and EfficientNet-B0 as the backbone networks, each patch is represented by a 512 and 1280-dimensional features vector, respectively. The number of \(K\) will vary depending from the number of tumour patches in their corresponding slide. Slide labels will be inherited into their corresponding patches. Thus, for all bags of instances \(\{B_1, B_2, ..., B_N\}\) with \(N\) as the number of slide, every \(x\) in \(B\) will have same label as their corresponding slide \(Y_n \in \{0,1\}, n=1...N\).\\
	
	We used attention mechanism as an aggregation layer as widely used in \cite{ilse2018attention,qiu2021attention,zhang2022dtfd}.  An attention layer in aggregation layer is defined as a weighted sum:
	\begin{equation}
		z = \sum_{k=1}^{K}a_k x_k,
	\end{equation}
	where 
	\begin{equation}
		a_k = \frac{\exp \{\text{w}^\intercal \, \text{tanh} \, (\text{V}x_{k}^{\intercal})  \}}
		{\sum_{j=1}^{K} \exp \{\text{w}^\intercal \, \text{tanh} \, (\text{V}x_{j}^{\intercal})\}}
	\end{equation}
	and \(\text{w} \in \mathbb{R}^{L\times 1}\) is weight vector of each instance, \(\text{V} \in \mathbb{R}^{L\times M}\). \(\text{tanh}\) is hyperbolic tangent function which is used in the first hidden layer. The second layer employs a softmax non-linearity function to ensure that the attention weights sum is equal to one. The attention weights can be interpreted as the relative importance of the instance. The higher the attention weight, the more important an instance in the final prediction score. Furthermore, this mechanism also creates a fully trainable aggregation layer. 
	
	We applied the ReLU non-linearity function for each fully connected layer except for the last of fully connected layers. Each output from the fully connected layer will go through the next fully connected layer and the aggregation layers. This model has three weighted sums \(z_1, z_2, z_3\) that are then accumulated in the last aggregation via the residual connection. The accumulation of weighted sum from each aggregation layer is then used in the last fully connected layers to predict label of the bag.
	
	We trained our model with learning rate of \(1\times{10}^{-4}\) and with Adam optimization algorithm. All the experiments were implemented on Python language and Pytorch framework. We conducted our experiment on workstation with Intel(R) i5-10500 CPU (3.1GHz), 64 GB RAM and single GPU NVIDIA RTX 3090.
	
	\subsection{Metrics for Evaluation}
	We use the accuracy score to measure our model performance. Due to the imbalanced class distribution in our dataset, we also use F1-score and  Area Under the Curve and Receiver Operating Characteristic (AUROC) curve. F1-score is defined as follow:
%	\begin{equation}
%		Accuracy = \frac{TP+TN}{TP+TN+FP+FN}
%	\end{equation}
	\begin{equation}
		F1 =  \frac{TP}{TP+\frac{1}{2}(FP+FN)}
	\end{equation}
	where TP, TN, FN, and FP are true positive, true negative, false negative and false positive, respectively. All the experiments were conducted on three-folds cross validation with stratified label. Due to the usage of cross-validation in our experiment, the average and standard deviation of all metrics were also reported.

	\section{Results}
	\subsection{Tissue Classification}
	Tissue classifier model achieved accuracy of 0.990 and F1-score of 0.970 on the test set. This model then used on the LMP1 cohort to stratify tumour patches. We extracted non-overlapping patches with the size of \(224\times 224\) at a 20x magnification level from LMP1 cohort. We conducted a segmentation on lower-resolution image to find foreground and background region in image. We use the segmentation result as a guidance to segregate patches with tissue and background patches. Only patches contains tissue were inferred in the tumour classification process.  
	
	\begin{figure}
		\centering
		\includegraphics[width=0.8\textwidth]{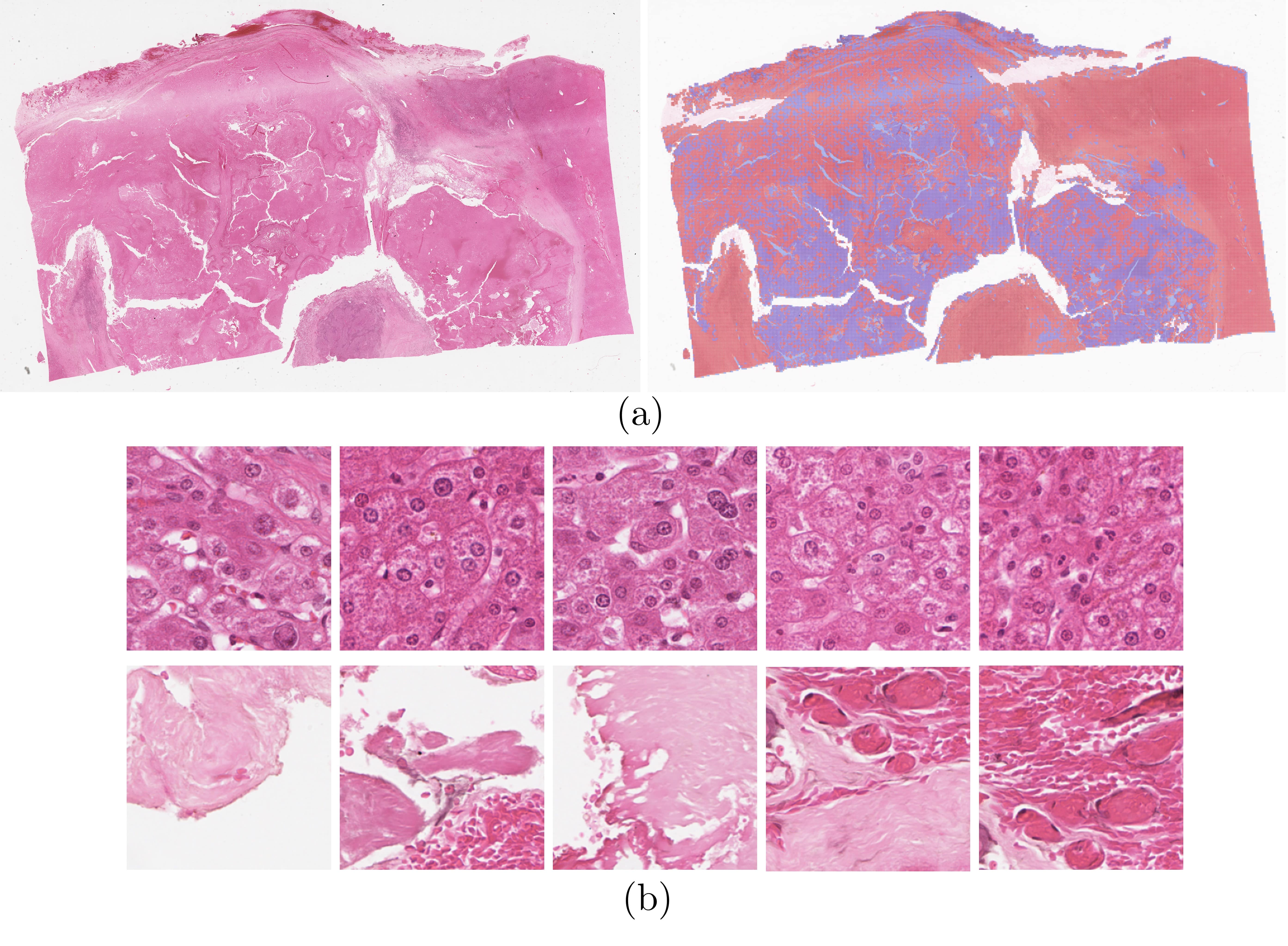}
		\caption{\textbf{Tissue Classification Result.} (a) Top left image is a thumbnail of one of slide in the dataset and top right image is the corresponding tumour region heatmap. (b) Samples of tumour patches is depicted in the top row and non-tumour region in the bottom row} \label{fig_tissue_classification}
	\end{figure}
	
	As can be seen in Figure \ref{fig_tissue_classification}, tumour patches were separated nicely in the WSIs. There was some minor misclassification in some slides due domain shift problem, but the rate was too low and could be ignored. Backbone networks encoded these tumour patches into vector representation and used them as bag of instances for LMP1 status prediction.

	\subsection{LMP1 Prediction}
	We argue that the tumour region is more important than any other tissue region for predicting LMP1. The non-tumour region may hinder the performance of LMP1 prediction and weighs irrelevant regions. To this end, we compared the performance of model with all tissue region as bag of instance vs model which only employs tumour patches as bags of instance. We also compared two backbone networks, namely ResNet-18 and EfficientNet-B0 in this experiment. The result of this experiment can be seen in Table \ref{tab:table_result1}. There were differences in performance between the same backbone networks. The model that trained with only the tumour region delivered better performance predicting LMP1 status than the model that trained with all tissue regions. Furthermore, EfficienetNet-B0 was better than ResNet-18 in terms of the backbone network. The best performance was achieved by using tumour regions as instances and EfficientNet-B0. This model achieved average accuracy, AUC and F1-score of 0.936, 0.995 and 0.862, respectively.
	\setlength{\tabcolsep}{4pt}
	\begin{table}%[ht]
		\begin{center}
			\caption{LMP1 Status Prediction Results}
			\label{tab:table_result1}
			\begin{tabular}{lllll}
				\hline\noalign{\smallskip}
				Tissue Region & Backbone Network & Accuracy & AUC & F1-score\\
				\noalign{\smallskip}
				\hline
				\noalign{\smallskip}
				All region & ResNet-18 & 0.748 \textpm 0.035 & 0.877 \textpm 0.013 & 0.310 \textpm 0.257 \\
				All region & EfficientNet-B0 & 0.926 \textpm 0.032	& 0.978	\textpm 0.014 & 0.825 \textpm 0.095 \\				
				Tumour region & ResNet-18 & 0.831 \textpm 0.082 & 0.954 \textpm 0.023 &	0.423 \textpm 0.364\\				
				Tumour region & EfficientNet-B0 & 0.936 \textpm 0.030 & 0.995 \textpm 0.002 &	0.862 \textpm 0.076 \\					
				\hline
			\end{tabular}
		\end{center}
	\end{table}
	\setlength{\tabcolsep}{1.4pt}

	We also compared our result with other known methods such as CLAM \cite{lu2021data} and MI-Net \cite{wang2018revisiting}. In the original pipeline, CLAM uses three sets of data, namely training, validation and testing sets. We modify the CLAM pipeline only to use the training and testing set to allow a fair comparison. Two aggregation methods were examined for MI-Net: the max and mean methods. As for the feature vectors in MI-Net, we use the same as our proposed model, which feature vectors from tumour patches encoded by EfficientNet-B0. The number of outputs of fully-connected layers in the MI-Net are 256, 128 and 64, respectively. MI-Net was trained with Adam optimizer and learning rate of \(1\times{10}^{-4}\). We also ensure CLAM and MI-Net use the same data as ours in the cross-validation training.		
	\setlength{\tabcolsep}{4pt}
	\begin{table}
		\begin{center}
			\caption{Performance Comparison of Our Model}
			\label{tab:table_comparison}
			\begin{tabular}{lllll}
				\hline\noalign{\smallskip}
				Model & Tissue Region & Accuracy & AUC & F1-score\\
				\noalign{\smallskip}
				\hline
				\noalign{\smallskip}
				CLAM \cite{lu2021data} & All region & 0.723 \textpm 0.022 & 0.548 \textpm 0.092 & - \\\hdashline \noalign{\smallskip}
				MI-Net Max \cite{wang2018revisiting} & All region & 	0.852  \textpm 0.011 & 0.700  \textpm 0.018 & 0.571 \textpm 0.037 \\ 					
				MI-Net Mean \cite{wang2018revisiting} & All region 	& 0.762 \textpm	0.002 & 0.541 \textpm 0.016 & 0.165 \textpm 0.076 \\ 						
				MI-Net Max \cite{wang2018revisiting} & Tumour region&	0.832 \textpm 0.031 & 0.661 \textpm	0.054 &	0.477 \textpm0.130		 \\ 
				MI-Net Mean \cite{wang2018revisiting} & Tumour region & 	0.767 \textpm0.006 & 0.559 \textpm0.041 & 0.207 \textpm0.136 \\\hdashline \noalign{\smallskip}						
				Our model & Tumour region & \textbf{0.936} \textpm 0.030 & \textbf{0.995} \textpm 0.002 &	\textbf{0.862} \textpm 0.076 \\
				\hline
				
			\end{tabular}
		\end{center}
	\end{table}
	\setlength{\tabcolsep}{1.4pt}		
			
	\begin{figure}%[ht]
		\centering
		\includegraphics[width=\textwidth]{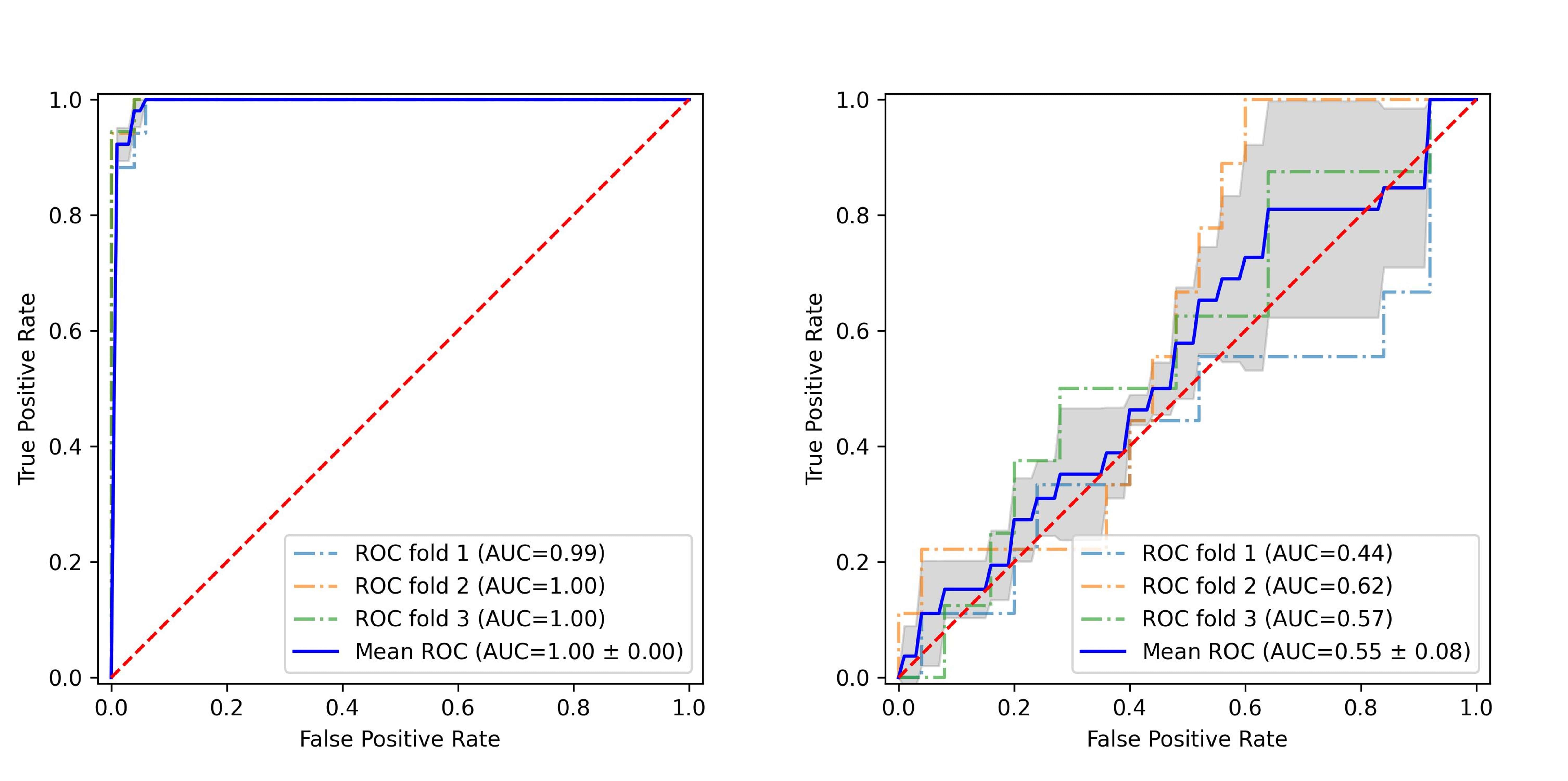}
		\caption{\textbf{AUROC curve} Left image is AUROC curve for our model and right image is AUROC curve for CLAM} \label{fig_roc}
	\end{figure}	
	Based on the performance comparison in Table \ref{tab:table_comparison} and AUROC in Figure \ref{fig_roc}, CLAM performed poorly on this particular task. The average accuracy score achieved was only 0.723, with average AUC of 0.548. On the other hand, MI-Net performed well in all experiments compared to the CLAM. Both max and mean aggregation methods surpass CLAM performance in accuracy and AUC. However, aggregation with the max method was better than the mean method in this case. This may occur because the mean method averages from many other irrelevant instances while the max method selects only relevant instances. Our model achieved the best accuracy, AUC and F1-score compared to other models. The second-best model was MI-Net with max-aggregation that trained on tumour regions of the tissue. Both of MI-Net and our models performed better when only tumour regions was used as bag of instances. This indicates that tumour region is more relevant to predicting LMP1 in our case.
	
	\begin{figure}
		\centering
		\includegraphics[width=0.8\textwidth]{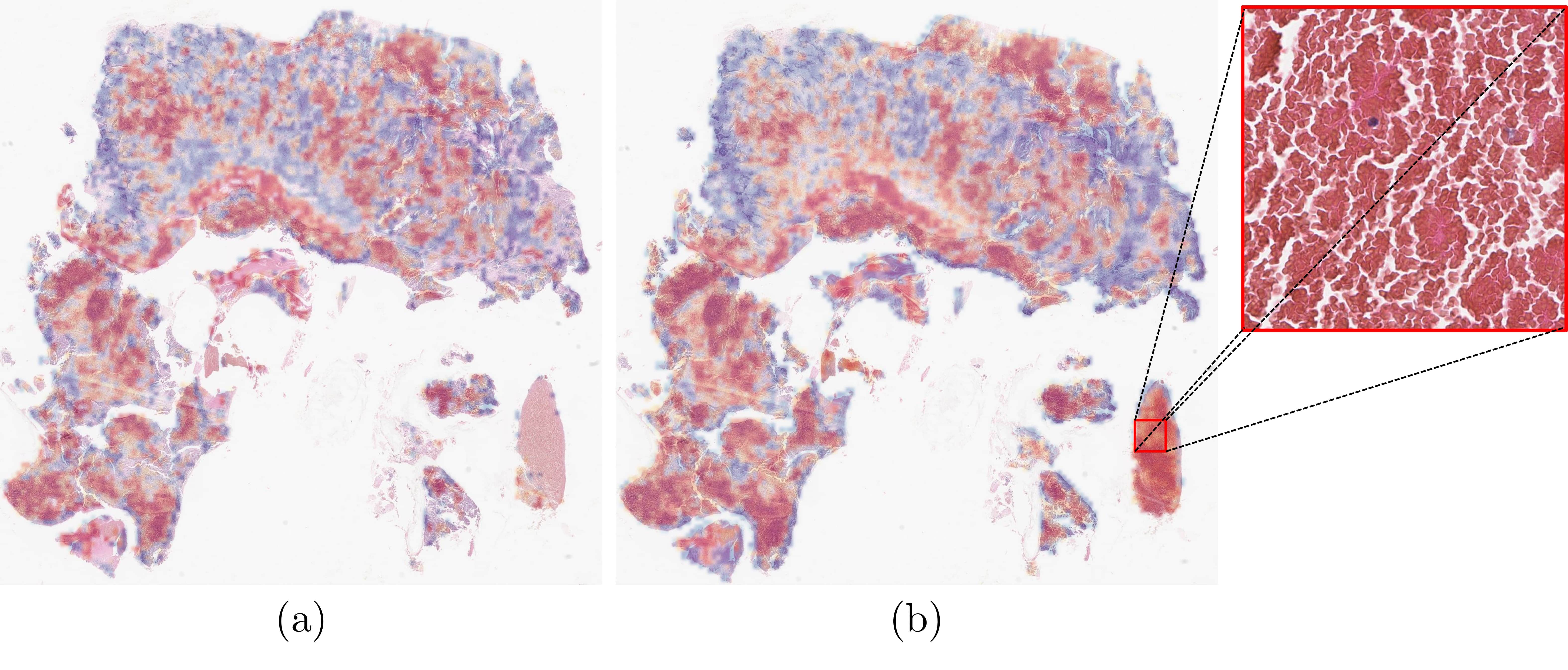}
		\caption{\textbf{Prediction comparison on different instance types.} LMP1 heatmap prediction with tumour patches as instances(a) and all patches as instances(b). Region with high attention score in(b) was blood cell, which irrelevant region regarding to LMP1.} \label{fig_compare_att}
	\end{figure}

	Utilising all regions in the tissue may generate an unmeaningful result. This example can is shown in Figure 4. There are two models which trained with a different formulation of the bag of instances. The right image is the attention heatmap of the model, which is trained on all regions of tissue, and the left image is trained on tumour regions. The model on the right side weighs blood cell regions with a high attention score, which is irrelevant in predicting LMP1  status.
	
	\subsection{Model Interpretability}	
\begin{figure}[ht]
		\centering
		\includegraphics[width=0.9\textwidth]{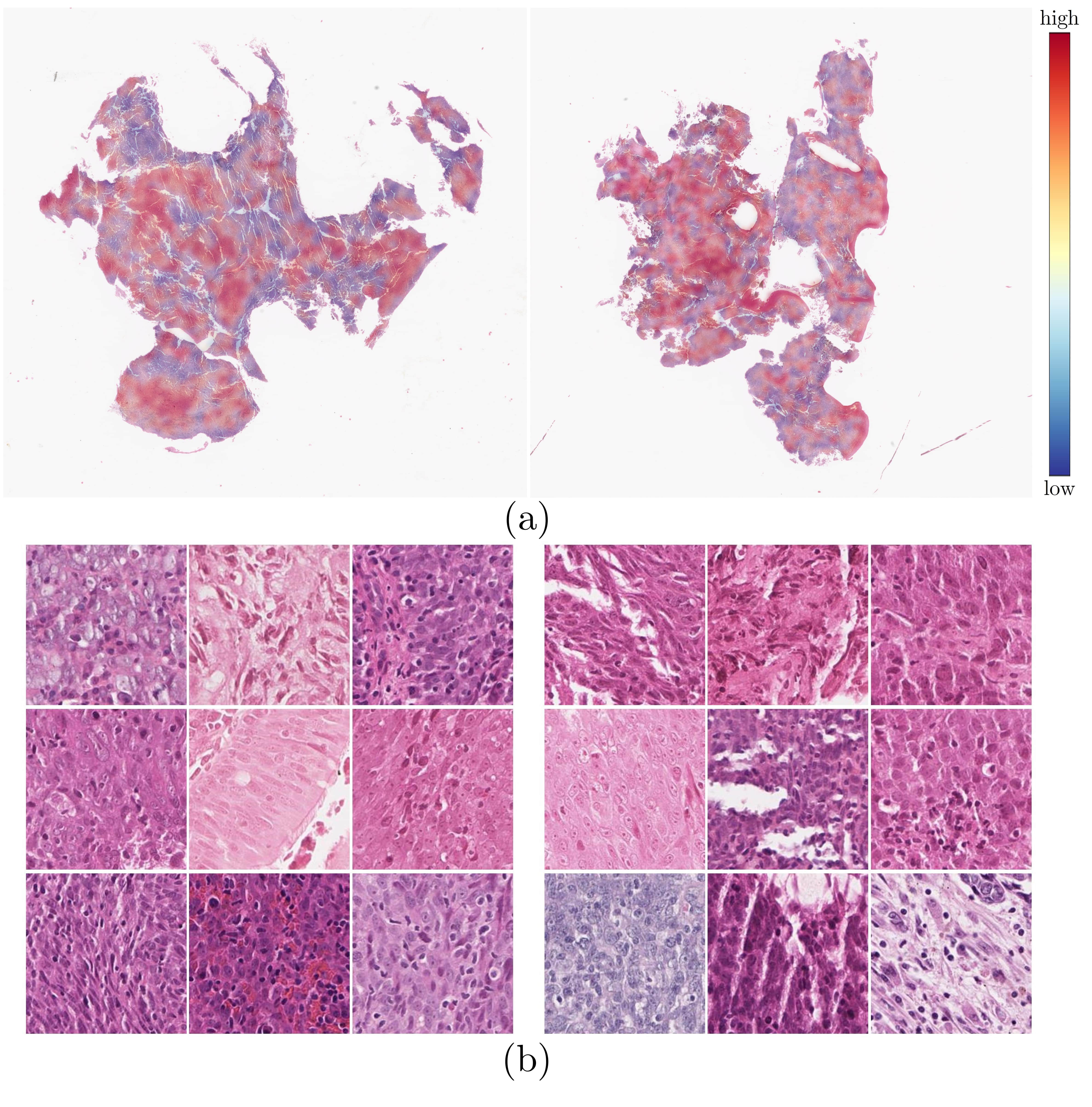}
		\caption{\textbf{Attention heatmap of LMP1 prediction model.} (a) is attention heatmap, the left image is LMP1 negative and the right image is LMP1 positive. (b) is randomly selected patches from dataset with highest attention score, left images is true negative cases and right images is true positive cases} \label{fig_attention_heatmap}
	\end{figure}

	The advantage of attention-based multiple instance learning is the interpretability of the model. We can inspect the relative importance among instances in regards to bag labels. Attention heatmaps from both of LMP1 negative and positive can be seen in Figure \ref{fig_attention_heatmap}. This figure also visualised the patches with top 10\% attention scores from both LMP1 classes. Based on the colour heterogeneity of patches, colour variation from the staining method was uncorrelated to LMP1 status. 	
	
	We also conducted a simple test to prove that the multiple attention layer chooses the most relevant instances in regards to the bag label. Currently, there is no well-defined explanation regarding LMP1 effect on tumour micro-environment and its morphology. Therefore, we use MNIST as an example. We trained the same MIL architecture as the LMP1 status prediction task on the MNIST dataset. We defined the task into a binary classification. Bags which contained at least one image of digit 1 were defined as the positive class, and bags without image of digit 1 were defined as the negative class. We generated 5000 bags with a balanced class distribution. Each bags contain ten images/instances. The number of digits 1 was randomly selected in the positive class. We trained our model with the same learning rate \(1\times{10}^{-4}\) and optimized it with the Adam algorithm. To encode the image digits in MNIST into a features vector, we flatten the image of 28x28 into a 1-dimensional array of 784 elements. We then changed the scale of the feature vector from 0-255 to 0-1.
	
	\begin{figure}[ht]
		\centering
		\includegraphics[width=\textwidth]{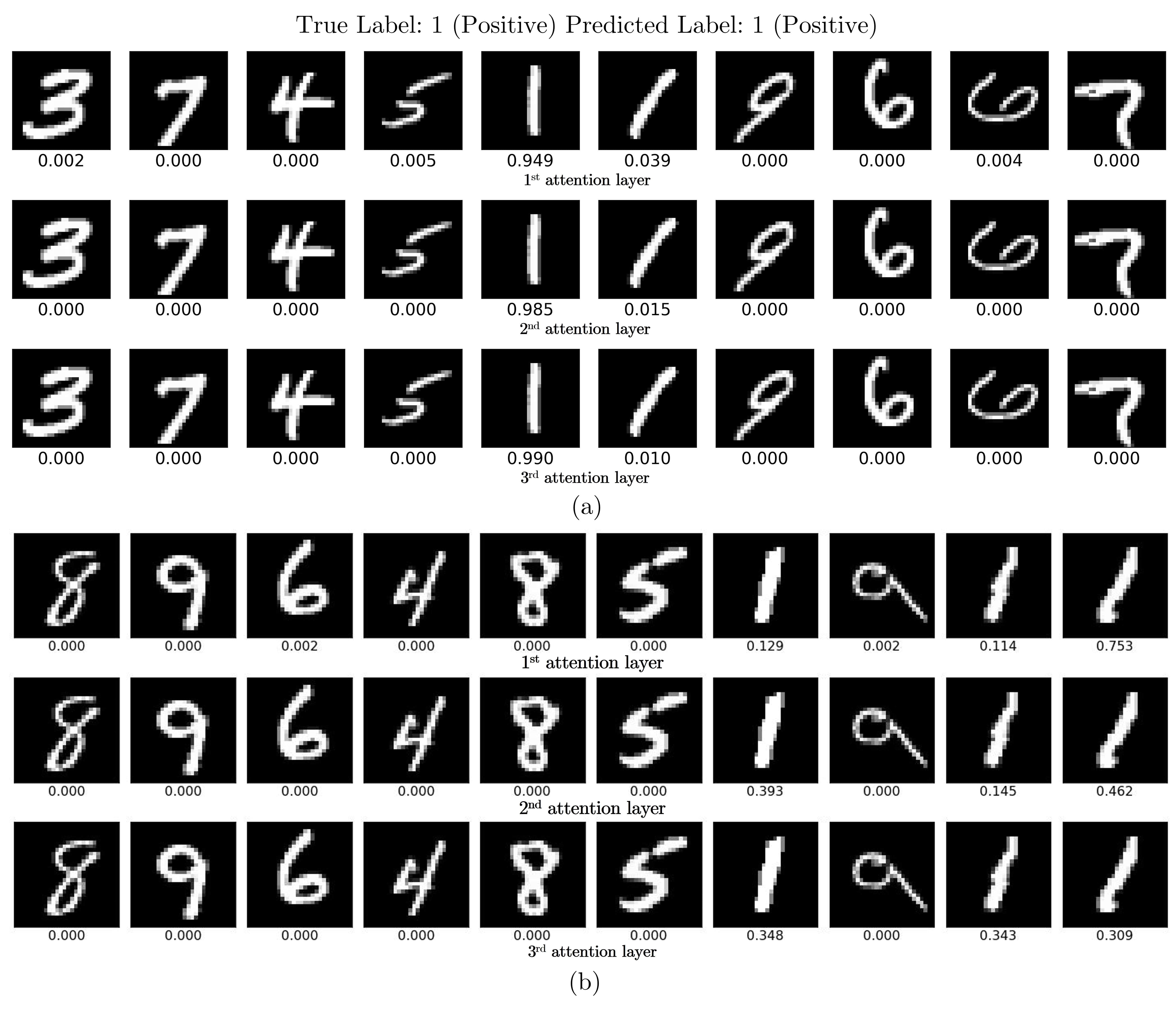}
		\caption{\textbf{Attention Scores from Three Aggregation Layers} These are two examples of bags in MNIST (a,b) with their corresponding attention scores in three aggregation layers. Number below each digits is the attention scores.} \label{fig_mnist}
	\end{figure}

	Figure \ref{fig_mnist} depicts an example of inferred results from a positive bag. Bag (a) contains two positive instances/two images of digit 1. These instances have the highest attention scores among other instances in the first aggregation layers. In the second attention layer, the attention scores for negative instances were reduced while the scores for positive instances were increased. This trend continues until the last aggregation layer. Thus all negative instances have zero value of attention. In another case, in bag (b), the attention scores in the last layers were averaged among the positive instances. This experiment proves that multiple attention-based of aggregation layers help the model to select more relevant instances.

	\section{Conclusion}
	In this paper, we examine the use of deep learning for LMP1 status prediction in NPC patients using the H\&E-stained WSIs. LMP1 data in this study was collected from a University Hospital with a total number of 101 cases. We proposed multi attention aggregation layer for MIL in the tumour region to predict LMP1 status. There was an increase in the model performance when using only tumour regions as instances. Despite the simplicity of our proposed method, it outperformed the other known MIL models. To the best of our knowledge, this is the first attempt to predict LMP1 status on NPC using deep learning. 
	
	\section{Acknowledgements}
	This study is fully supported by a PhD scholarship to the first author funded by Indonesia Endowment Fund for Education (LPDP), Ministry of Finance, Republic of Indonesia under grant number Ref: S-575/LPDP.4/2020.

	\clearpage
	\bibliographystyle{splncs04}
	\bibliography{egbib}
\end{document}